\documentclass[11pt]{article}

\usepackage[margin=1in]{geometry}
\usepackage[T1]{fontenc}
\usepackage[utf8]{inputenc}
\usepackage{lmodern}
\usepackage{microtype}
\usepackage{enumitem}
\usepackage[hidelinks]{hyperref}
\usepackage{parskip}

\title{RIACT: A Responsible AI System for Personalized Study Habit Tracking and Early Burnout Signal Detection in University Students}
\author{Ria Sidhu\\ \small \texttt{riasidhu02@gmail.com}}
\date{}

\begin{document}

\maketitle

\begin{abstract}
Student burnout is highly prevalent in higher education, with reported rates ranging from 12\% to over 70\% and consistently exceeding those of the working population --- yet it is typically identified only retrospectively, after academic decline has already occurred. A contributing factor is that students have little structured visibility into their own study behaviour, and existing productivity tools record activity without interpreting it. This paper presents RIACT (Record, Insight, Analyze, Coach, Track), a web-based application that combines structured study session logging with a hybrid AI architecture to surface personalized insights and early burnout signals. Students log sessions by location and time; the system computes net focus time by accounting for breaks, detects burnout signals through transparent, deterministic rules operating on week-over-week behavioural comparisons, and uses a large language model --- constrained to a fixed output schema --- to contextualize patterns and generate personalized recommendations. The design embeds responsible AI principles throughout: warnings are governed by auditable rules rather than model judgement, all output is framed as an observation rather than a diagnosis and data collection is limited to self-logged behavioural fields. We describe the system's design rationale, situate it within the literature on student burnout and explainable AI in education and propose an evaluation framework for validating its behavioural signals against established burnout instruments.
\end{abstract}

\section{Introduction}

University students face mounting academic pressure, and burnout has emerged as one of the most significant threats to student wellbeing and performance. Research indicates that burnout prevalence among university students ranges from 12\% to over 70\%, with studies consistently finding that rates exceed those of the working population (Olson et al., 2025). Despite its prevalence, burnout is most commonly identified only after it has already taken hold, at which point intervention is remedial rather than preventive.

A contributing factor is that students have little visibility into their own study behaviour. Without data on where and when they study most effectively, students repeat unproductive habits and fail to recognize warning signs in their own patterns. Existing tools such as time trackers and productivity apps record activity, but offer no personalized analysis tied to individual learning outcomes. Generic study advice, however well-intentioned, cannot substitute for insight grounded in a student's own history.

Part of the difficulty is that self-monitoring without structure is unreliable. Students asked to estimate how long they studied, how often they took breaks or when their focus was strongest are reconstructing from memory and those reconstructions are systematically distorted --- long sessions feel productive regardless of how fragmented they were, and gradual changes in behaviour are precisely the kind that memory smooths over. A student whose sessions have been shrinking for three weeks rarely experiences that as a trend; each individual session simply feels like a busy day. The behavioural signals that precede burnout are, in other words, visible in the data but largely invisible to the person generating it. This is why structured logging matters: it externalizes the record, making week-over-week comparison possible in a way that introspection is not.

Several commercial tools exist for tracking time and productivity. Toggl Track\footnote{\url{https://toggl.com/track}} and RescueTime\footnote{\url{https://www.rescuetime.com}} are among the most widely used, allowing users to log time spent on tasks and generate reports on productivity patterns. Notion\footnote{\url{https://www.notion.com}} and similar productivity platforms offer flexible note-taking and task management with some habit-tracking capabilities. While these tools are useful for general time management, they share a common limitation in the context of student wellbeing: they record activity without interpreting it. A student who logs four hours of study time in RescueTime receives a report of four hours, but no insight into whether those hours were focused or fragmented, whether the location was conducive to concentration or whether the pattern across the week signals approaching exhaustion. The tools are descriptive, not analytical.

None of these platforms integrate burnout detection. None personalize recommendations based on the individual's behavioural history. Lastly, none are designed specifically for the rhythms of academic life, where location, time of day and goal-based deadlines interact in ways that generic productivity frameworks do not capture. This gap motivated the design of RIACT.

This paper presents RIACT (Record, Insight, Analyze, Coach, Track), a web-based application that addresses this gap. RIACT enables students to log study sessions by location and time, automatically calculates net focus by accounting for breaks and after a minimum of three sessions applies AI to surface personalized patterns, detect early burnout signals and deliver actionable recommendations. The system is designed with responsible AI principles throughout; observations are surfaced instead of diagnoses, while keeping the student in control of any decisions that follow.

The remainder of this paper is organized as follows. Section~\ref{sec:related} reviews related work in study tracking, learning analytics and AI-based student wellbeing interventions. Section~\ref{sec:design} describes the system architecture and design of RIACT. Section~\ref{sec:discussion} discusses the implications of this approach, its limitations and directions of future work. Section~\ref{sec:conclusion} concludes the paper.

\section{Related Work}
\label{sec:related}

\subsection{Student Burnout in Higher Education}
\label{sec:burnout-lit}

Student burnout is a well-documented phenomenon in higher education research. Maslach and Jackson (1981) established the foundational framework for measuring burnout across three dimensions: emotional exhaustion, cynicism and reduced academic efficacy --- a model later adapted specifically for university students by Schaufeli et al. (2002), whose cross-national study established that burnout is measurable in student populations using the same dimensional structure as occupational burnout. In the student adaptation, emotional exhaustion reflects fatigue from study demands, cynicism reflects mental distancing from one's studies and reduced academic efficacy reflects a diminished sense of competence and accomplishment.

Subsequent research has confirmed that burnout prevalence among students is alarmingly high, with reported rates ranging from 12\% to over 70\% depending on the population and methodology, consistently exceeding rates observed in the working population (Olson et al., 2025). In a recent cross-sectional study of 947 German university students across five academic fields, Olson et al. (2025) found that 73.2\% of students were moderately to highly stressed and 28.2\% reported frequent burnout symptoms in at least one dimension --- and notably, that burnout was not concentrated in the medical fields that dominate literature; informatics and engineering students showed the highest stress and burnout levels. Studies of undergraduate populations elsewhere report similar prevalence (Li et al., 2021), and systematic reviews confirm the breadth of the problem across disciplines and institutions (Almutairi et al., 2022).

Two findings from this literature are particularly relevant to RIACT's design. First, burnout develops gradually and is typically identified late --- the instruments used to measure it are administered retrospectively, capturing burnout that has already manifested. Second, Salmela-Aro and Read (2017) demonstrate that burnout and engagement exist on a continuum, with identifiable profiles between full engagement and full burnout, suggesting that behavioural decline is detectable before burnout becomes acute. This is specifically the window in which RIACT is designed to operate: its signals --- declining session length, rising break frequency, late-night displacement --- are behavioural proxies for the erosion of engagement that the literature identifies as burnout's precursor.

\subsection{AI in Educational Contexts}

The integration of AI tools into higher education has accelerated in recent years. Nguyen et al. (2024), studying ChatGPT adoption among university students, found the tool widely used for learning support, information search, idea generation and assignment completion, with many students regarding it as reliable for academic tasks. But the same study surfaced consistent concerns from both students and faculty: overreliance may erode independent thinking and critical evaluation, alongside risks of academic dishonesty and reduced creativity. These concerns share a root cause --- the AI is open-ended, general-purpose and disconnected from the student's actual learning context, so its output substitutes for the student's own work rather than informing it.

RIACT takes a deliberately different approach to AI in the student's life. Rather than providing an open-ended interface that can generate academic content, it constrains AI to a single domain --- analysis of the student's logged behaviour --- where the AI's output cannot substitute for studying but only inform decisions about it. The risks identified by Nguyen et al. (2024) are structurally excluded rather than policed.

Khosravi et al. (2022) propose the XAI-ED framework for explainable AI in education, arguing that concerns about fairness, accountability, transparency and ethics of AI-supported educational interventions are best addressed by systems that produce transparent explanations for the decisions they make. They emphasize that explainability in education has distinctive needs --- the stakeholders include learners who must be able to trust and act on the system's output. RIACT aligns with these principles at the architectural level: its most sensitive function, burnout warning, is produced by transparent deterministic rules whose triggering signals are shown to the user (Section~\ref{sec:signals}), rather than by an opaque model score.

\subsection{Ethics and Responsibility in AI for Education}

Schiff (2022), analyzing 24 national AI policy strategies, found that the use of AI in education is largely absent from policy conversations and that the ethical implications of educational AI receive scant attention even in documents that otherwise discuss AI ethics prominently. The gap matters because AI systems in education interact with populations, including students under academic stress --- where poorly designed interventions can cause real harm, and because effective policy and careful ethics are, as Schiff argues, inextricably linked.

In the absence of settled policy guidance, responsibility falls to system designers. RIACT's response is a set of self-imposed constraints, described fully in Section~\ref{sec:responsible}: observations rather than diagnoses, deterministic rules rather than model judgement for safety-relevant decisions, minimal data collection and links to professional resources without clinical claims. These choices position RIACT within an emerging standard of care for AI tools deployed in wellbeing-adjacent educational contexts --- one that treats the vulnerability of the user population as a design input rather than an afterthought.

\section{System Design}
\label{sec:design}

RIACT (Record, Insight, Analyze, Coach, Track) is a web-based application built to help university students understand their study habits through structured session logging and AI-generated insights.

\subsection{Architecture}

RIACT is built on Next.js with a TypeScript frontend styled using Tailwind CSS. Authentication and data persistence are handled by Supabase, which enforces row-level security on every database table; ensuring users can only ever access their own records, a critical requirement given the sensitive nature of personal productivity and wellbeing data. The application is deployed on Vercel. AI features are powered by the OpenAI API using the GPT-4o-mini model, selected for its balance of reasoning capability and response latency within a serverless environment.

\subsection{Session Tracking}

Users initiate a study session by selecting a location and projected end time. A live timer runs for the duration of the session and automatically pauses when a break is recorded. Net study time (total elapsed time minus break duration) is stored alongside location, start time, end time and break frequency. Raw study time alone is a poor proxy for focus; a three-hour session with six breaks reflects a fundamentally different cognitive state than an uninterrupted hour. By capturing break patterns explicitly, RIACT builds a richer signal than duration alone.

\subsection{Data Model}
\label{sec:datamodel}

RIACT's data model consists of four core entities. A \emph{session} records the location, start time, end time, projected end time, total elapsed minutes and computed net study minutes. A \emph{break} belongs to a session and records its own start time, end time and duration; net study time is derived by subtracting accumulated break duration from total session time. A \emph{goal} specifies a target number of study hours over a daily or weekly timeframe, optionally scoped to a specific location and carries an active/inactive state. A \emph{location} is a user-defined named place, persisted so that repeat entries accumulate under a consistent label --- a prerequisite for any per-location analysis. Every table is scoped to the authenticated user through row-level security, so no query can cross user boundaries even in the event of client-side error.

This schema is deliberately minimal. RIACT collects no biometric data, no device telemetry and no content of what is being studied; only where, when and how continuously. This restraint is itself a design decision; the system demonstrates that meaningful behavioural inference is possible from a small, transparent set of self-logged fields without the invasive data collection common to commercial productivity software.

\subsection{User Flow}

A student's first interaction with RIACT begins with account creation via email and password authentication. On first login, the home screen presents the student's active goals, which are initially empty, and a prompt to begin a study session. The student may optionally create goals first, specifying a target number of study hours on a daily or weekly basis, optionally tied to a specific location (for example, six hours per week at the campus library).

To begin a session, a student selects a location, either from a saved list of previously used locations or by entering a new one, and sets a projected end time representing their intention for the session. This projected end time serves a dual purpose; it anchors the student's commitment before the session begins, and the gap between projected and actual end times later becomes a behavioural signal in its own right, since consistently ending sessions early can indicate declining engagement.

During an active session, a live timer counts upward. When the student takes a break, they record it with a single action; the timer pauses and resumes when the break ends. On ending the session, the student is shown a review screen summarizing the session (location, start and end times, breaks taken and computed net study time), with full editing support before the record is saved. This review step is deliberate: it gives the student a moment of reflection on the session that just occurred and it allows correction of logging errors (such as forgetting a break) that would otherwise silently degrade the quality of the data underlying all subsequent AI analysis.

Once three sessions have been logged, the AI features become available. The insights view surfaces detected patterns and any burnout signals; the dashboard aggregates statistics across days and weeks, including per-location breakdowns; the coach provides a conversational interface over the student's own history; and the weekly plan generates a day-by-day study schedule grounded in the student's demonstrated patterns and active goals. From this point forward, every additional session refines the picture the system holds of the student's habits, and the recommendations sharpen accordingly.

\subsection{Burnout Signal Detection}
\label{sec:signals}

Burnout detection in RIACT is deliberately \emph{not} delegated to the language model. Instead, signals are computed by a deterministic, rule-based algorithm operating on a rolling comparison of the most recent seven days of sessions against the seven days prior. Four signals are evaluated:

\begin{enumerate}
    \item \textbf{Declining session length} --- triggered when average net study time in the last seven days falls below 85\% of the preceding week's average.
    \item \textbf{Rising break frequency} --- triggered when the average number of breaks per session increases by more than 25\% week over week.
    \item \textbf{Late-night studying} --- triggered when more than two sessions in the last seven days begin at or after 10PM, a pattern associated with cramming and sleep displacement.
    \item \textbf{Declining goal completion} --- triggered when fewer than half of users' active goals were met in the last seven days.
\end{enumerate}

If any signal fires, a burnout warning is surfaced along with the specific signals that triggered it. This architecture has two advantages. First, it is fully transparent and reproducible: a user (or a researcher) can trace exactly why a warning appeared, consistent with the explainability principles discussed in Section~\ref{sec:related}. Second, it isolates a safety-relevant function from the nondeterminism of a language model --- the LLM interprets and contextualizes patterns, but the decision to warn is made by auditable rules.

\subsection{AI Layer}
\label{sec:ailayer}

AI features activate after a minimum of three logged sessions. This threshold is intentional; drawing inferences from one or two sessions risks producing recommendations that reflect noise rather than genuine patterns, which could mislead users at a stage when trust in the system is still being established. When triggered, a server-side API route fetches the user's last 30 days of sessions and active goals from Supabase using a Bearer-token authenticated client, so that row-level security is enforced even on server-side queries. The session data is reduced to a minimal structured payload --- location, start and end times, net and total minutes and break records --- and passed to GPT-4o-mini with a prompt that constrains the response to a fixed JSON schema: a one-sentence pattern summary, a categorical burnout risk level, a list of contributing signals and exactly three short recommendations. Constraining the output format serves both reliability (the frontend can parse responses deterministically) and restraint (the model cannot produce open-ended commentary in a context where measured, bounded output is appropriate).

Three AI-driven features are built on this foundation:

\begin{itemize}
    \item \textbf{Pattern Recognition:} identifies which location and time of day produces the highest net study time relative to break frequency, giving students actionable information they cannot easily derive themselves.
    \item \textbf{Risk Contextualization:} alongside the rule-based detector described in Section~\ref{sec:signals}, the model provides an independent, holistic assessment of burnout risk from the full session history --- the deterministic rules govern when a warning is surfaced, while the model's assessment enriches the insights view with interpretation.
    \item \textbf{Personalized Recommendations:} generates specific, actionable guidance based on the user's own historical patterns rather than generic advice.
\end{itemize}

For the AI study coach, the same session context is passed conversationally, allowing users to ask open-ended questions about their habits. The weekly study plan feature returns structured JSON which the frontend renders as a day-by-day schedule.

\subsection{Responsible AI Design}
\label{sec:responsible}

RIACT is designed with deliberate constraints around AI output, reflecting the sensitivity of its domain: a tool that comments on a student's behavioural patterns during periods of academic stress operates closer to wellbeing territory than a typical productivity application, and its design must account for that.

The most important constraint is that burnout detection surfaces signals and warnings but never tells a user they \emph{are} burned out; that determination is left to the user. All AI output is framed as observation rather than diagnosis --- the system reports that average session length is declining or that late-night sessions are increasing, but does not translate those observations into claims about the user's mental state. This distinction is more than tonal. Burnout, as discussed in Section~\ref{sec:related}, is a construct with clinical dimensions measured by validated instruments; a consumer application inferring it from behavioural proxies has no legitimate basis for diagnostic language, and using such language could cause a stressed student to either over-identify with a label or dismiss a genuinely useful warning.

When burnout signals are detected, the system links to mental health resources but makes no clinical claims and recommends no treatment. The boundary is explicit: RIACT can observe behaviour and prompt reflection; anything beyond that belongs to the student and, where appropriate, to qualified professionals.

Several structural decisions reinforce these constraints. As described in Section~\ref{sec:signals}, the decision to surface a burnout warning is made by transparent, deterministic rules rather than by the language model, ensuring the most sensitive function of the system is auditable and reproducible. The prompt design described in Section~\ref{sec:ailayer} binds the model's output to a fixed schema, preventing open-ended commentary in context where measured output is appropriate. And the minimal data model described in Section~\ref{sec:datamodel} limits collection to self-logged behavioural fields --- no biometrics, no device surveillance, no content of what is studied --- so the system's insight is bounded by what the student has chosen to share.

Finally, the user retains control at every point of the data lifecycle: sessions are reviewed and editable before saving, goals can be deactivated or deleted, and AI features are advisory rather than gatekeeping --- no functionality is withheld or altered based on the AI's assessment of the user. The design reflects an explicit commitment to human-in-the-loop decision making, in a context where users may already be under significant academic stress.

\section{Discussion}
\label{sec:discussion}

\subsection{Why Personalization Matters}

Existing productivity tools, such as Toggl and RescueTime, provide aggregate time tracking but offer no personalized analysis tied to learning outcomes. Generic study advice --- ``study in the morning,'' ``avoid distractions'' --- fails to account for individual variation in peak focus windows, preferred environments and workload patterns. This variation is not noise to be averaged away; it is the signal. The literature reviewed in Section~\ref{sec:related} found meaningful differences in stress and burnout across academic disciplines within a single university, and there is no reason to expect less variation at the level of individual habits. Advice calibrated to the average student is, almost by definition, miscalibrated for most actual students.

RIACT takes a different approach: rather than prescribing behaviour based on population-level averages, it surfaces patterns derived exclusively from the individual user's own data. A recommendation to study Tuesday mornings at a specific location carries more actionable weight than any generalized tip because it is grounded in that student's demonstrated history. It is also more falsifiable in a useful sense --- the student can follow it and observe, through continued logging, whether the pattern holds. This creates a feedback loop generic advice cannot offer: recommendations that improve as the data accumulates, and a student who becomes progressively more informed about their own behaviour rather than progressively more dependent on external guidance.

\subsection{Early Burnout Detection as a Preventive Tool}
\label{sec:early}

Student burnout is typically identified retrospectively; after academic performance has already declined, after withdrawal from activities, after the student themselves recognizes something is wrong. The instruments that measure burnout, however well validated, are administered at discrete points and capture a state that has already developed. By the time burnout is visible through these channels, intervention is remedial rather than preventive.

RIACT's burnout detection is designed to operate earlier in this trajectory. The behavioural shifts it monitors --- declining session length, increasing break frequency, late-night cramming --- are continuous, passive byproducts of normal app use rather than assessments the student must remember to take. The student who would never think to complete a burnout inventory in week six of the semester is nonetheless generating, through ordinary logging, exactly the data in which week six's warning signs appear. This is the core preventive claim: not that RIACT detects burnout better than validated instruments, but that it watches continuously in the long stretches between the moments anyone would think to measure.

This does not replace clinical support, and RIACT makes no claim that it does. Rather, it surfaces a prompt for self-reflection at a point when the student still has meaningful agency to respond.

\subsection{Design Tradeoffs}
\label{sec:tradeoffs}

Several design decisions in RIACT reflect deliberate tradeoffs rather than default choices:

\textbf{General-purpose model versus custom model.} RIACT uses GPT-4o-mini, a general-purpose language model, rather than a model trained specifically on learning analytics data. A purpose-built model could in principle produce better-calibrated insights, but would require training data that does not yet exist at this scale, and would sacrifice the conversational flexibility that powers the study coach. The hybrid architecture mitigates the main risk of this choice: because burnout warnings are governed by deterministic rules rather than the model, the least reliable component of the system is never responsible for the most safety-relevant decision.

\textbf{Location-based rather than subject-based tracking.} RIACT organizes sessions around \emph{where} studying happens rather than \emph{what} is being studied. This was a deliberate scoping decision. Subject-based tracking demands more logging effort per session and produces data that fragments quickly across courses, while location and time-of-day patterns remain comparable across a student's entire history. The cost is that RIACT cannot currently distinguish between subjects that may demand different study styles --- a limitation discussed in Section~\ref{sec:limitations}.

\textbf{The three-session threshold.} Requiring three sessions before AI activation is a compromise between statistical caution and user retention. A higher threshold would produce more reliable early insights but risks users abandoning the app before ever seeing its core value. Three sessions is early enough to demonstrate value within a student's first week of use, while still being defensible as a minimum basis for pattern commentary --- and the system's language is calibrated accordingly, presenting early insights as provisional observations rather than conclusions.

\textbf{Manual logging rather than automatic tracking.} RIACT requires students to actively start sessions and record breaks, rather than inferring activity from device usage as tools like RescueTime do. Manual logging introduces self-report bias and friction. But it also keeps the student consciously engaged with their own behaviour --- the act of logging is itself a reflective intervention --- and it avoids the privacy costs of passive surveillance, consistent with the minimal data model described in Section~\ref{sec:datamodel}.

\subsection{Limitations}
\label{sec:limitations}

RIACT has not yet been evaluated with a real user cohort, and this is its most significant limitation. The burnout detection signals are theoretically grounded --- each corresponds to behavioural patterns documented in the burnout literature --- but the specific thresholds (a 15\% decline in session length, a 25\% rise in break frequency, more than two late-night sessions per week) have not been validated against established clinical measures, such as the Maslach Burnout Inventory. It is possible that these thresholds are too sensitive, producing warnings that users learn to ignore, or too conservative, missing early decline. Only empirical calibration against self-reported outcomes can resolve this.

The system also inherits the limitations of self-reported data. Students who log inconsistently, forget breaks or backfill sessions from memory will receive insights built on degraded data, and the system currently has no mechanism for detecting logging quality. Relatedly, the population most at risk of burnout may be precisely the population least likely to maintain consistent logging --- a disengaged student stops logging before they stop studying --- meaning the system could lose visibility exactly when its warnings would matter most.

Several scoping decisions carry costs. Location-based tracking cannot distinguish between subjects with different cognitive demands. The seven-day comparison windows make the detector responsive to recent change but blind to slow, semester-long decline. And GPT-4o-mini, while capable, is a general-purpose model whose interpretations, though bounded by the prompt schema, are not guaranteed to be consistent across identical inputs.

Finally, novelty effects cannot be ruled out: any measured engagement with a new self-tracking tool tends to decline after initial enthusiasm and RIACT's long-term value depends on whether the insight loop is compelling enough to sustain logging beyond the first weeks.

\subsection{Evaluation Considerations}

Evaluating a system like RIACT requires separating three questions that are easily conflated: whether students will use it, whether its inferences are accurate and whether it changes outcomes.

The first is a question of sustained engagement. The relevant measure is not downloads or initial sign-ups but logging persistence --- the proportion of participants still recording sessions consistently after four, eight or twelve weeks. Because self-tracking tools reliably show engagement decay after initial novelty, short evaluation windows would overstate RIACT's viability; an evaluation must run long enough for novelty effects to dissipate.

The second is a question of signal validity. Each burnout signal makes an implicit empirical claim --- that its behavioural pattern precedes or accompanies genuine burnout. Testing this requires pairing the system's continuous behavioural data with a validated instrument such as the MBI-SS administered at regular intervals. The analysis is then straightforward in principle: do participants whose logged data triggers RIACT's detector score higher on emotional exhaustion and cynicism than those whose data does not? Signal-level analysis is also possible --- late-night displacement may prove more predictive than break frequency or vice versa --- allowing the detector to be refined rather than merely validated or rejected wholesale.

The third is a question of impact, and it is the hardest. Even if students use RIACT and its signals are valid, the system only matters if surfaced insights change behaviour --- whether students actually shift their study locations and times toward their detected peaks and whether warned students adjust their patterns in the weeks following a warning. A controlled comparison would assign some participants the full system and others a logging-only version with AI features disabled, isolating the contribution of the insight layer from the reflective effect of logging itself --- an effect the design deliberately exploits, as noted in Section~\ref{sec:tradeoffs}, but which must be separated analytically from the value of the AI.

Two confounds deserve particular attention in any such design. Academic workload varies dramatically across a term and behavioural signals will fluctuate with exam periods regardless of well-being; evaluation windows must therefore span full academic cycles rather than sampling within them. And self-selection is unavoidable --- students who volunteer for a study-tracking trial are unlikely to be representative of the disengaged students the system most aims to reach, a limitation that should temper generalization from any pilot cohort.

\subsection{Future Work}

The natural next step is a structured pilot study with a cohort of university students over a full academic term. Such a study would recruit participants across disciplines --- the literature reviewed in Section~\ref{sec:related} suggests burnout prevalence varies meaningfully by field of study --- and measure whether students using RIACT demonstrate improved study consistency, earlier burnout awareness, or better goal completion relative to a control group. Administering a validated instrument such as the MBI-SS at regular intervals would allow the rule-based signals to be tested directly: do weeks in which RIACT's detector fires correspond to elevated scores on validated burnout dimensions? Each signal threshold could then be calibrated empirically rather than heuristically.

Beyond validation, several system extensions follow naturally. Logging-quality detection could flag periods of sparse or irregular data and qualify the confidence of insights accordingly. Longer comparison windows could complement the current seven-day detector to capture semester-scale trends. And integration with class schedules --- which RIACT already stores --- opens the possibility of contextualizing study patterns against academic workload, distinguishing a student who studies less because demands have eased from one who studies less because they are disengaging.

\section{Conclusion}
\label{sec:conclusion}

Student burnout and ineffective study habits represent a persistent and underaddressed challenge in higher education. While the problem is well documented, students have lacked practical tools that translate their own behavioural data into personalized, actionable guidance. RIACT addresses this gap by combining structured session logging with AI-driven pattern recognition, burnout detection and coaching --- all grounded in the individual student's history rather than generalized advice.

This paper has described the design and architecture of RIACT and argued that personalized, data-driven feedback represents a more effective approach to study habit intervention than existing generic tools. While empirical validation remains future work, the system is designed with responsible AI principles that make it appropriate for deployment with a vulnerable population. RIACT is openly accessible and represents a foundation for future research into AI-assisted student wellbeing.

\section*{References}
\begin{list}{}{\setlength{\itemindent}{-0.35in}\setlength{\leftmargin}{0.35in}\setlength{\itemsep}{4pt}}

\item Almutairi, H., Alsubaiei, A., Abduljawad, S., Alshatti, A., Fekih-Romdhane, F., Husni, M., \& Jahrami, H. (2022). Prevalence of burnout in medical students: A systematic review and meta-analysis. \emph{International Journal of Social Psychiatry, 68}(6), 1157--1170. \url{https://doi.org/10.1177/00207640221106691}

\item Khosravi, H., Buckingham Shum, S., Chen, G., Conati, C., Tsai, Y.-S., Kay, J., Knight, S., Martinez-Maldonado, R., Sadiq, S., \& Ga\v{s}evi\'{c}, D. (2022). Explainable Artificial Intelligence in education. \emph{Computers and Education: Artificial Intelligence, 3}, 100074. \url{https://doi.org/10.1016/j.caeai.2022.100074}

\item Li, Y., Cao, L., Liu, J., Zhang, T., Yang, Y., Shi, W., \& Wei, Y. (2021). The prevalence and associated factors of burnout among undergraduates in a university. \emph{Medicine, 100}(27), e26589. \url{https://doi.org/10.1097/MD.0000000000026589}

\item Maslach, C., \& Jackson, S. E. (1981). The measurement of experienced burnout. \emph{Journal of Organizational Behavior, 2}(2), 99--113. \url{https://doi.org/10.1002/job.4030020205}

\item Nguyen, T. N. T., Lai, N. V., \& Nguyen, Q. T. (2024). Artificial Intelligence (AI) in education: A case study on ChatGPT's influence on student learning behaviors. \emph{Educational Process: International Journal, 13}(2), 105--121. \url{https://doi.org/10.22521/edupij.2024.132.7}

\item Olson, N., Oberhoffer-Fritz, R., Reiner, B., \& Schulz, T. (2025). Stress, student burnout and study engagement -- a cross-sectional comparison of university students of different academic subjects. \emph{BMC Psychology, 13}, 293. \url{https://doi.org/10.1186/s40359-025-02602-6}

\item Salmela-Aro, K., \& Read, S. (2017). Study engagement and burnout profiles among Finnish higher education students. \emph{Burnout Research, 7}, 21--28. \url{https://doi.org/10.1016/j.burn.2017.11.001}

\item Schaufeli, W. B., Mart\'{i}nez, I. M., Marques Pinto, A., Salanova, M., \& Bakker, A. B. (2002). Burnout and engagement in university students: A cross-national study. \emph{Journal of Cross-Cultural Psychology, 33}(5), 464--481. \url{https://doi.org/10.1177/0022022102033005003}

\item Schiff, D. (2022). Education for AI, not AI for education: The role of education and ethics in national AI policy strategies. \emph{International Journal of Artificial Intelligence in Education, 32}, 527--563. \url{https://doi.org/10.1007/s40593-021-00270-2}

\end{list}

\end{document}